\documentclass[11pt]{article}

\usepackage[preprint]{acl}

\usepackage{times}
\usepackage{latexsym}
\usepackage{amssymb}

\usepackage[T1]{fontenc}

\usepackage[utf8]{inputenc}

\usepackage{microtype}

\usepackage{inconsolata}

\usepackage{graphicx}
\usepackage{booktabs}
\usepackage{mathtools}
\usepackage{multirow}

\title{Beyond Reference-Based Evaluation: Reward Models for Meta-Evaluation of Grammatical Error Correction}

\author{
 \textbf{Ruotian Wu\textsuperscript{1,2}},
 \textbf{Bill E. Johnson\textsuperscript{3}},
 \textbf{Gene Saunders\textsuperscript{3}},\\
 \textbf{Osama Hamzeh\textsuperscript{3}},
 \textbf{Ankit Vadehra\textsuperscript{1,2,3}},
 \textbf{Pascal Poupart\textsuperscript{1,2}},
\\
\\
 \textsuperscript{1}University of Waterloo,
 \textsuperscript{2}Vector Institute,
 \textsuperscript{3}Scribendi Inc.
\\
 \small{
   \textbf{Correspondence:} \href{mailto:r82wu@uwaterloo.ca}{r82wu@uwaterloo.ca}
 }
}

\begin{document}
\maketitle
\begin{abstract}
Reference-based metrics for Grammatical Error Correction (GEC) such as M$^2$ and ERRANT assume that the reference set enumerates all valid edits, and therefore often penalize corrections that are grammatical and meaning-preserving but phrased differently. We introduce \textbf{RM-EVAL}, a reward model trained on human preference data from SEEDA, as a \emph{reference-free} meta-evaluator that predicts human-like quality judgments at both full-sequence and partial-sequence levels. Beyond evaluation, we show that the same reward model can be used as a learning signal to \emph{improve} GEC generation via \textbf{Reward-Guided Text Generation (RGTG)}, which keeps a base GEC model frozen and performs online, reward-driven decoding. Across SEEDA, RM-EVAL achieves strong agreement with human rankings, and RGTG yields consistent gains in reward and external validation, demonstrating a unified framework for both assessing and enhancing GEC systems without relying on gold references.
\end{abstract}

\vspace{-2mm}
\section{Introduction}
\vspace{-2mm}
Evaluation remains one of the central challenges in Grammatical Error Correction (GEC). Metrics such as M$^2$ \citep{dahlmeier2012m2} and ERRANT \citep{bryant2017automatic} are widely used, yet they depend on fixed references that capture only a limited set of acceptable corrections. Consequently, any deviation—even if grammatically or semantically valid—may be penalized. This reliance on references limits generalization and fails to align well with human judgments.

Recent work in reward modeling and reinforcement learning from human feedback (RLHF) has demonstrated the potential of learned evaluators to approximate human preferences \citep{ouyang2022training}. In this paper, we propose \textbf{RM-EVAL}, a reward model trained on the SEEDA dataset—a large-scale collection of human evaluation data for GEC systems. RM-EVAL provides a reference-free evaluation framework that captures human-like judgments GEC quality.

This work makes the following contributions:
\vspace{-2mm}
\begin{itemize}
    \item We propose a \textbf{reward-model–based evaluation framework} (\textbf{RM-EVAL}) as a reference-free alternative that achieves competitive or superior correlation with human preferences compared to conventional metrics, and performs on par with state-of-the-art (paid closed) LLM-based evaluation methods, while being much smaller, free and open source.
    \vspace{-2mm}
    \item We further demonstrate that RM-EVAL can be used to improve \emph{any} text-generation GEC model (including LLM-based correctors) via \textbf{Reward-Guided Text Generation (RGTG)}, an online RLHF-style decoding procedure that \emph{keeps the base GEC model frozen} while steering outputs using the learned reward signal.
\end{itemize}

\vspace{-2mm}
\section{Related Work}
\vspace{-2mm}


\paragraph{Reference-based Evaluation.}
Standard GEC metrics such as M$^2$ and ERRANT compute overlap between system edits and reference edits, effectively penalizing any non‑reference corrections. This reliance on limited, often incomplete reference sets leads to three key shortcomings: (1) the unrealistic assumption of reference completeness, (2) inconsistent scores across different reference annotations, and (3) poor alignment with human judgments of adequacy and fluency \citep{chollampatt2018reassessment}. These limitations motivate the need for reference‑free evaluation approaches.
\vspace{-2mm}
\paragraph{Meta-evaluation Benchmarks.}
Recent benchmarks such as SEEDA \citep{SEEDA} have provided large-scale human ratings of GEC outputs, enabling systematic meta-evaluation of automatic metrics. However, these benchmarks are still limited by the metrics they test rather than providing new evaluation paradigms.

\vspace{-2mm}
\paragraph{Reward Models for Evaluation.}
Reward modeling has gained traction as a mechanism to train evaluators that generalize human preferences across diverse tasks, including summarization and translation \citep{stiennon2020learning}. To our knowledge, this work is the first to explore reward models for meta-evaluation of GEC.
\vspace{-2mm}
\paragraph{Reward-Guided Text Generation (RGTG).}
RGTG is an online, reward-driven generation paradigm that improves outputs at inference time using a learned reward function, without updating the base generator’s parameters. Rather than performing standard RL fine-tuning of the policy, RGTG keeps the underlying model frozen and steers generation using reward feedback during decoding, enabling lightweight alignment and easier reuse across generators and domains \citep{rashidcritical}.

\vspace{-2mm}
\section{Methodology}
\vspace{-2mm}
\subsection{Problem Setup and Data}
We consider GEC as a conditional text generation task: given a source sentence $x$, a system produces a correction $y$.
Our supervision for evaluation is \emph{preference data} of the form $(x, y^{(a)}, y^{(b)}, p)$, where $p \in \{a,b\}$ indicates which output humans prefer under criteria such as grammaticality, fluency, and meaning preservation.

A crucial requirement is that the compared candidates are both \emph{valid alternative corrections of the same source sentence}.
For example, it is appropriate to construct a pair from two GEC system outputs for the same $x$ with a human preference label, or from a source--target pair $(x, y^\star)$ where $y^\star$ is explicitly judged as preferred for that $x$.
In contrast, it is generally problematic to treat an arbitrary gold reference $r$ as ``preferred'' over a system output $y$ \emph{unless the preference label was obtained by directly comparing $y$ and $r$}, because we otherwise do not know which one humans would prefer when both may be acceptable.  In this work, we use SEEDA \citep{SEEDA}, which provides human judgments enabling reliable preference supervision for training a learned evaluator.
\vspace{-2mm}
\subsection{RM-EVAL: Reward Model for Reference-Free Evaluation}
We train a reward model $R_\theta(x, y)$ that assigns a scalar quality score to a correction $y$ conditioned on source $x$.
Given a labeled preference pair $(x, y^{+}, y^{-})$, we optimize a standard pairwise preference objective:
\begin{equation}
\mathcal{L}(\theta) = - \log \sigma\big(R_\theta(x, y^{+}) - R_\theta(x, y^{-})\big),
\end{equation}
where $\sigma(\cdot)$ is the logistic sigmoid. The trained reward model can score a full correction sequence. Furthermore, with the temporal difference loss in Eq.~\ref{eq:constraint_loss}, the reward model can also be applied to partial sequences \cite{rashidtowards} to support fine-grained analysis (e.g., scoring prefixes during generation).
\begin{equation} \scalebox{0.92}{$\displaystyle
\label{eq:constraint_loss}
            \mathcal{L}(\theta) = \frac{1}{2}\left[R_{\theta}(x, y_{1:i}^{+}) - \max_{y_{i+1}^{+}} \:R_{\theta}(x, y_{1:i+1}^{+})\right]^2   
            $}
\end{equation}

\vspace{-2mm}
\subsection{Applying RGTG to GEC with a Frozen Generator}
Beyond evaluation, we leverage the trained reward model as a learning signal for improving GEC outputs via \textbf{Reward-Guided Text Generation (RGTG)} \citep{rashidtowards}.
RGTG is an inference-time alignment framework that steers generation using a learned reward function, without updating the parameters of the underlying generator.

Formally, let $G_\phi$ denote a pretrained or instruction-tuned GEC model that maps a source sentence $x$ to a correction $y$.
In contrast to standard RLHF approaches that fine-tune $G_\phi$ using policy optimization, RGTG keeps $G_\phi$ \emph{frozen} and incorporates the reward model $R_\theta(x, y)$ directly into the decoding process.
This design significantly reduces computational cost and avoids instability associated with policy updates, while still enabling alignment with human preferences.

\vspace{-2mm}
\paragraph{Reward-Guided Decoding Objective.}
At decoding step $t$, given a partial hypothesis $y_{<t}$, the generator proposes a candidate set $\mathcal{C}_t$ of next-token or short-span continuations sampled from $G_\phi$.
Each candidate $y_{t} \in \mathcal{C}_t$ yields an extended hypothesis $y_{<t} \oplus y_t$ where $\oplus$ represents concatenation, then each hypothesis is evaluated using the reward model.
RGTG selects the continuation that maximizes a reward-augmented decoding objective:
\begin{equation}
\scalebox{0.95}{$\displaystyle
\begin{aligned}
c^* = \arg\max_{y_{t} \in \mathcal{C}_t}
\Big(
& \log P_\phi(y_{t} \mid x, y_{<t}) \\
& + \beta \, \exp(R_\theta(x, y_{<t} \oplus y_{t}))
\Big)
\end{aligned}
$}
\end{equation}
where $\beta$ controls the trade-off between model likelihood and reward guidance.

Because $R_\theta$ can be applied to partial sequences, reward feedback is incorporated incrementally throughout generation, rather than only after producing a complete correction.
This enables fine-grained control over decoding and allows the reward model to influence local generation decisions.

Applied to GEC, RGTG directly optimizes for human-aligned correction quality—capturing grammaticality, fluency, and meaning preservation—without relying on reference edits or additional supervised data.
Moreover, since the generator remains unchanged, the same reward model can be reused to steer different GEC systems, making RGTG a flexible and cost-effective mechanism for improving correction quality at inference time.

\vspace{-2mm}
\section{Experiments}
\vspace{-2mm}
\subsection{Setup}
We evaluate RM-EVAL on the SEEDA benchmark following the meta-evaluation protocol described in \citet{SEEDA}. Baselines include M$^2$, ERRANT, existing GEC-specific metrics such as GLEU, and LLM based methods described in \citet{kobayashi-etal-2024-large}. We measure the performance by accuracy and Kendall’s rank correlation ($\tau$) \citep{kendall1938new} with human judgments at sentence level. All the reward models are trained by fine-tuning Llama3.2-1B\footnote{\href{https://huggingface.co/meta-llama/Llama-3.2-1B-Instruct}{meta-llama/Llama-3.2-1B-Instruct}} or Qwen2.5-1.5B\footnote{\href{https://huggingface.co/Qwen/Qwen2.5-1.5B-Instruct}{Qwen/Qwen2.5-1.5B-Instruct}}.

\vspace{-2mm}
\subsection{Results}
\subsubsection{Reward Model Evaluation on SEEDA}
We first evaluate \textbf{RM-EVAL} through 5-fold cross-validation on the SEEDA dataset. Each fold consists of a non-overlapping subset of human-annotated preference pairs. For each fold, we train on the remaining folds and test on the current fold.
Performance is reported in terms of (1) \textbf{pairwise accuracy}—the percentage of correctly predicted human-preferred outputs, and (2) \textbf{Kendall's correlation} between human rankings and rankings based on model-predicted reward scores.


\begin{table}[h!]
\centering
\resizebox{1.0\linewidth}{!}{
\begin{tabular}{llcccccc}
\toprule
\textbf{Model} & \textbf{Metric} & fold1 & fold2 & fold3 & fold4 & fold5 & \textbf{Avg.} \\
\midrule
\multirow{2}{*}{Llama} 
 & Accuracy    & 0.82 & 0.80 & 0.81 & 0.79 & 0.81 & \textbf{0.81} \\
 & Correlation & 0.57 & 0.58 & 0.57 & 0.59 & 0.62 & \textbf{0.59} \\
\midrule
\multirow{2}{*}{Qwen} 
 & Accuracy    & 0.87 & 0.84 & 0.85 & 0.82 & 0.82 & \textbf{0.84} \\
 & Correlation & 0.63 & 0.62 & 0.59 & 0.64 & 0.61 & \textbf{0.62} \\
\bottomrule
\end{tabular}
}
\caption{Five-fold cross-validation of RM-EVAL on SEEDA.  
Correlation denotes Kendall's rank correlation between human rankings and reward model predictions.}
\label{tab:rmseeda}
\end{table}

\begin{table}[t!]
\centering
\resizebox{0.70\linewidth}{!}{
\setlength{\tabcolsep}{6pt}
\begin{tabular}{lcc}
\toprule
\textbf{Metric} & \textbf{Acc} & \textbf{$\tau$} \\
\midrule
M$^2$ & 0.50 & 0.17 \\
ERRANT & 0.47 & 0.13 \\
GoToScorer & 0.50 & 0.01 \\
PT-M$^2$ & 0.53 & 0.18 \\
GLEU & 0.61 & 0.23 \\
SOME & 0.76 & 0.53 \\
IMPARA & 0.74 & 0.50 \\
Llama3 & 0.53 & 0.06 \\
GPT-3.5 & 0.61 & 0.25 \\
GPT-4 & 0.80 & 0.60 \\
GPT-5 & \textbf{0.83} & \textbf{0.63} \\
\midrule
{RM-EVAL (Llama)} & 0.81 & 0.59 \\
\textbf{RM-EVAL (Qwen)} & \textbf{0.84} & \textbf{0.62} \\
\bottomrule
\end{tabular}
}
\caption{
Sentence-level evaluation on SEEDA\_S with fluency correction. 
Reported metrics are Accuracy (Acc) and Kendall’s rank correlation coefficient ($\tau$). 
Higher values indicate stronger agreement with human judgments.
}
\label{tab:seeda_sentence}
\end{table}
The Qwen-based reward model achieves an average accuracy of 0.84 and a correlation of 0.62 with human ratings—comparable to SOTA LLM based metrics shown in Table \ref{tab:seeda_sentence}. These results demonstrate that RM-EVAL effectively captures human preferences for grammaticality and fluency without requiring annotations.  Furthermore RM-EVAL is small, open-source and free in comparison to GPT-5, which is a large, paid and closed model.

\vspace{-2mm}
\subsubsection{Reward-Guided Text Generation (RGTG) on SEEDA}
We next apply the trained reward model to Reward-Guided Text Generation (RGTG) - an online RLHF-style generation method.  
The base generator is Llama3-1B, and we compare variants fine-tuned with different alignment methods: ARGS \citep{khanov2024argsalignmentrewardguidedsearch}, PARGS \citep{rashidcritical}, and FaRMA \citep{rashidtowards}. We refer to the base instruction-tuned Llama3 model without other enhancement as the SFT baseline.

\begin{table}[h!]
\centering
\resizebox{1.0\linewidth}{!}{
\begin{tabular}{lcc}
\toprule
\textbf{Method} & \textbf{Reward Mean $\uparrow$} & \textbf{Std. Error} \\
\midrule
SFT & 3.02 & 0.15 \\
ARGS & 2.98 & 0.17 \\
PARGS & 2.96 & 0.18 \\
FaRMA & \textbf{3.53} & \textbf{0.05} \\
\bottomrule
\end{tabular}
}
\caption{Average reward scores assigned by RM-EVAL across RGTG variants on SEEDA.  
FaRMA achieves the highest mean reward and lowest variance, indicating the strongest alignment with human preferences.}
\label{tab:rgtg}
\end{table}

In Table \ref{tab:rgtg}, the FaRMA-aligned model obtains the highest mean reward (3.53), surpassing both ARGS and PARGS.  
To assess statistical reliability, we conduct pairwise \textbf{Wilcoxon signed-rank tests} on reward scores between model variants (Table~\ref{tab:significance}).
FaRMA shows statistically significant improvements over SFT ($p=0.030$) and ARGS ($p=0.038$), while the improvement over PARGS is positive but not statistically significant ($p=0.054$).
No statistically significant differences are observed among the remaining baseline comparisons.

\begin{table}[h!]
\centering
\begin{tabular}{lcccc}
\toprule
\textbf{Comparison} & \textbf{p-value} & \textbf{Sig.} \\
\midrule
FaRMA vs PARGS &  0.054 & -- \\
FaRMA vs SFT &  \textbf{0.030} & \checkmark \\
FaRMA vs ARGS &  \textbf{0.038} & \checkmark \\
\bottomrule
\end{tabular}
\caption{Pairwise Wilcoxon signed-rank tests on RGTG reward scores.
$\checkmark$ indicates statistical significance at $p<0.05$.}
\label{tab:significance}
\end{table}
\vspace{-2mm}
\subsubsection{GPT Evaluation as External Validator}
To further validate reward alignment, we conduct pairwise preference comparisons using GPT-5 as an external evaluator.  

\begin{table}[h!]
\centering
\begin{tabular}{l l c}
\toprule
\textbf{Method 1} & \textbf{Method 2} & \textbf{Win Rate (\%)} \\
\midrule
FaRMA & Base & 64 \\
FaRMA & ARGS & 68 \\
FaRMA & PARGS & 63 \\
\bottomrule
\end{tabular}
\caption{GPT-5 evaluation of pairwise win rates between model variants.}
\label{tab:gpt4}
\end{table}
\vspace{-2mm}
GPT-5 evaluation corroborates the reward model findings: FaRMA dominates in 68\% and 63\% of pairwise comparisons against ARGS and PARGS, respectively (Table \ref{tab:gpt4}).  
These results collectively suggest that RM-EVAL provides a reliable training signal for aligning generation models with human preferences.

\vspace{-2mm}
\section{Discussion}
\vspace{-2mm}
\paragraph{Reward Models as Meta-Evaluators.}
Our findings demonstrate that reward models can serve as reliable, reference-free meta-evaluation tools for GEC.  
Unlike traditional metrics such as M$^2$ or ERRANT that penalize valid non-reference corrections, RM-EVAL learns directly from human preferences and can generalize beyond specific edit patterns.  
The achieved human correlation of 0.62 matching the performance of GPT5—suggests that learned evaluators are approaching the ceiling of reference-based performance, while offering better interpretability in terms of preference modeling.
\vspace{-2mm}
\paragraph{Bridging Evaluation and Generation.}
An important implication of this work is the ability to use a single model both as an evaluator and as a learning signal for generation.  
When used as a reward function, RM-EVAL not only ranks system outputs consistently with human judgment but also guides models such as Llama3 toward more human-aligned behavior.
The FaRMA results (Tables~\ref{tab:rgtg}–\ref{tab:gpt4}) demonstrate that reinforcement learning guided by RM-EVAL yields tangible improvements, both in intrinsic reward scores and in GPT-based external evaluations.
\vspace{-2mm}
\paragraph{Generalization and Extensibility.}
Although our experiments focus on English GEC, the methodology is general and can be extended to other languages or text-editing tasks such as style transfer and paraphrasing.  
Since RM-EVAL is trained on preference pairs rather than explicit corrections, the framework is language-agnostic and amenable to multilingual scaling.
\vspace{-2mm}
\paragraph{Future Work.}
Future directions include expanding RM-EVAL into a multi-dimensional evaluator that independently scores fluency, adequacy, and grammaticality.
We also plan to explore multi-reward optimization for RLHF, where GEC models are optimized jointly for grammaticality and semantic preservation.  
Finally, integrating human-in-the-loop updates could further enhance the robustness and fairness of the reward model across diverse learner populations.
\vspace{-2mm}
\section{Conclusion}
\vspace{-2mm}
We presented \textbf{RM-EVAL}, a reward-model–based framework for meta-evaluation of grammatical error correction (GEC) systems.  
Unlike traditional reference-dependent metrics such as M$^2$ or ERRANT, RM-EVAL evaluates system outputs directly through learned human preferences, offering a scalable and reference-free approach to GEC evaluation.  
Empirical analyses on the SEEDA benchmark show that RM-EVAL achieves high correlation with human judgments and performs competitively with state-of-the-art metrics.  
When employed as a reward function for RGTG, the model further enhances system performance—particularly in the FaRMA variant—demonstrating the dual utility of reward models as both evaluators and optimization objectives.

\vspace{-2mm}
\section*{Limitations}
RM-EVAL relies on human preference annotations from the SEEDA benchmark for training. Although SEEDA provides high-quality human judgments, it covers only a specific dataset and evaluation setting. Consequently, the reward model may inherit biases introduced during the annotation process and reflect preferences specific to the annotator population. In scenarios involving different learner populations, domains, or correction styles, the reward model may not reliably capture human preferences. In such cases, additional preference annotations may be necessary to recalibrate the model and ensure robust evaluation performance.

\bibliography{custom}

\appendix
\section{Training Details}
All experiments are run on a server with NVIDIA RTX6000 GPUs (24GB VRAM) and NVIDIA A40 GPUs(40GB VRAM). We use CUDA Toolkit version 12.4 and PyTorch 2.5.1 framework. We train two versions of RM-EVAL on the SEEDA dataset, utilize the TRL library to accelerate the training process. We report the training parameters:

\vspace{5mm}

\begin{tabular}{ccc}
    \toprule & Parameters& Value\\
    \midrule
    \multirow{5}{*}[-0ex]{Llama} 
    & LR & 5e-6\\
    & Batch size & 16 \\
    & Gradient acc.\ steps & 16 \\
    & DeepSpeed Zero stage & 2 \\
    & Max. sequence length & 512 \\
    & $\beta$ (RGTG) & 0.5 \\
    \bottomrule
\end{tabular}

\vspace{5mm}

\begin{tabular}{ccc}
    \toprule & Parameters& Value\\
    \midrule
    \multirow{5}{*}[-0ex]{Qwen} 
    & LR & 5e-6\\
    & Batch size & 8\\
    & Gradient acc.\ steps & 8\\
    & DeepSpeed Zero stage & 2 \\
    & Max. sequence length & 512 \\
    \bottomrule
\end{tabular}


\end{document}